\documentclass[conference]{IEEEtran}
\IEEEoverridecommandlockouts

\usepackage{cite}
\usepackage{amsmath,amssymb,amsfonts}
\usepackage{graphicx}
\usepackage{textcomp}
\usepackage{xcolor}
\usepackage{booktabs}
\usepackage{multirow}
\usepackage{array}
\usepackage{stfloats}
\usepackage{url}

\def\BibTeX{{\rm B\kern-.05em{\sc i\kern-.025em b}\kern-.08em
    T\kern-.1667em\lower.7ex\hbox{E}\kern-.125emX}}

\begin{document}

\title{HybridDualPathNet: Cross-Fusion Attention of EfficientNetV2
and Swin Transformer for Brain MRI Classification of
Alzheimer's and Parkinson's Disease}

\author{
\IEEEauthorblockN{Jarpula Nirjala}
\IEEEauthorblockA{
\textit{Department of Electronics and Communication Engineering}\\
\textit{Indian Institute of Information Technology, Surat, India}\\
Email: nirjala8462@gmail.com
}
\and
\IEEEauthorblockN{Rahul Patel}
\IEEEauthorblockA{
\textit{Department of Electronics and Communication Engineering}\\
\textit{Indian Institute of Information Technology, Surat, India}\\
Email: rahulpatelanuppur@gmail.com
}
}
\maketitle

\begin{abstract}
Separating Alzheimer's Disease (AD), Parkinson's Disease (PD), and
Healthy Controls (HC) on a single structural MRI is a genuinely
awkward problem. The disease signatures overlap on the cortex, PD scans
are always in short supply, and---the part most papers quietly
skip---how the data is partitioned tends to decide the headline number
more than the network does. We study \textbf{HybridDualPathNet}, which
pairs EfficientNetV2-S (local texture) with Swin-Tiny (global context)
and fuses them through a Cross-Fusion Attention (CFA) unit that
decides, scan by scan, how much of each stream to trust, and we run it
through two experiments. The first is the conventional one: train and
test on the Kaggle APD-3-Class set alone under an image-level split,
where the model reaches \textbf{98.25\%} accuracy. The second is
stricter: one model trained and tested on Kaggle \emph{and} ADNI
together, 17{,}859 slices, under a per-source patient-wise split so no
subject's slices land on both sides. There the same network scores
\textbf{73.4\%} accuracy and \textbf{0.880} macro AUC, and splitting the
result by source exposes the real story---\textbf{91.2\%} on Kaggle-style
scans but only \textbf{59.2\%} on ADNI. Temperature scaling pulls
expected calibration error from 0.130 to 0.015, split-conformal
prediction gives 89.5\% coverage, and multi-method saliency confirms
the network looks at brain tissue rather than skull. The 25-point gap
between the two experiments is, honestly, the most useful result here.
\end{abstract}

\begin{IEEEkeywords}
Alzheimer's disease, Parkinson's disease, brain MRI, data leakage,
patient-wise split, cross-dataset generalization, ADNI, cross-fusion
attention, calibration, Grad-CAM++.
\end{IEEEkeywords}

\section{Introduction}
Neurodegenerative disease keeps adding to the load on health systems
everywhere. Alzheimer's is now diagnosed in upward of 55 million
people~\cite{b1}, Parkinson's in roughly ten million more~\cite{b2},
and both totals keep rising as the population ages. A T1-weighted
structural scan is usually the first thing a neurologist reaches for:
it is cheap, carries no ionizing dose, and tracks morphological change
well enough to be worth reading. The catch is that AD and PD leave
structural marks that look a lot alike, and radiologists themselves
part ways on the borderline cases~\cite{b3}. An automated reader that
could pre-screen these scans would take real pressure off overstretched
clinics, and deep networks routinely post accuracies in the high
nineties on public benchmarks.

We were, frankly, suspicious of those numbers, and this paper grew out
of that suspicion. Two habits in the literature quietly inflate
reported accuracy. First, the image-level split: MRI contains many
adjacent slices per patient, and if a random split scatters them across
train and test, a model can pass simply by recognizing a brain it has
already seen. Second, single-dataset evaluation, which never asks
whether a model tuned to one scanner survives contact with another. So
we report two experiments back to back---the ordinary Kaggle-only setup
first, then a leakage-controlled two-cohort setup---using the same
network for both, so the reader can see exactly how much the protocol,
not the model, is worth. Three familiar technical gaps sit underneath
all of this.

\textbf{Gap-1 (Architecture).} CNNs read fine local texture well but
miss how distant regions relate; Vision Transformers capture that
global context but blur local detail. No prior work on this benchmark
fuses both through a mechanism that adapts the blend per scan.

\textbf{Gap-2 (Data).} PD is badly under-represented---only 906 slices
on Kaggle, and just 967 in our combined set against thousands of AD and
CN. Plain oversampling invites memorisation; undersampling throws away
signal.

\textbf{Gap-3 (Calibration).} Accuracy and F1 are usually where the
reporting stops. For screening that is not enough: the dangerous case
is a network that is wrong while sounding certain, so we treat
calibrated probabilities as part of what the model has to deliver.

Our contributions are as follows.
\begin{itemize}
  \item The \textbf{CFA} fusion unit, whose per-sample, per-dimension
        gate mixes the convolutional and transformer streams through
        8-head attention in place of a static concatenation, letting
        the network lean on local hippocampal texture for AD-like scans
        and on global context for PD-like ones without any hand-set
        rule, at a cost of under a million extra parameters.
  \item A complete \textbf{single-dataset (Kaggle) study}---training,
        component ablation, a comparison against eight published
        methods, calibration, and Grad-CAM++ saliency---reaching
        98.25\% accuracy, reported first as the conventional
        image-level baseline that most prior work on this benchmark is
        measured against.
  \item A \textbf{leakage-controlled, two-cohort study} on Kaggle and
        ADNI together (17{,}859 slices), split per source and by patient
        so that no subject's slices cross partitions, with results
        reported \emph{by source} so the cross-scanner domain gap is
        measured rather than quietly averaged away.
  \item A direct \textbf{comparison} of the two experiments showing the
        same network---identical weights, loss, and training
        recipe---drops from 98.25\% to 73.4\% once leakage is removed,
        the clearest evidence here of how much single-dataset,
        image-level evaluation inflates the accuracies routinely
        reported on this task.
\end{itemize}

\section{Related Work}
For most of the past decade convolutional networks set the pace here.
ResNet~\cite{b5}, DenseNet, and the EfficientNet family~\cite{b6} land
somewhere around 89--94\% on AD/PD data~\cite{b_pradhan}, and
EfficientNetV2~\cite{b_efficientv2} tightened the speed--accuracy
balance with its fused-MBConv stages. Their shared weakness is
structural: convolutions relate nearby pixels well but struggle across
a whole slice. Vision Transformers~\cite{b7} and the Swin
Transformer~\cite{b8} close that gap with self-attention, though pure
transformers are data-hungry and wobble on the small cohorts typical of
neuroimaging---G\"uven~\cite{b_guven} still reached the mid-nineties
with one. Hybrids that join a CNN and a transformer have become popular
for exactly this reason: Hussain \textit{et al.}~\cite{b_hussain} report
about 97\% with a CNN--Swin combination, and Velu and
Jaisankar~\cite{b_velu} 97.4\% with an HDNN on APD-3-Class. Almost all
of these, however, glue the two feature streams with fixed weights and
report a single accuracy on a single dataset. The recurring problem is
not the architecture but the protocol---imbalance is patched rather than
measured, calibration is ignored, and the patient-versus-image split
often goes unstated. Guo \textit{et al.}~\cite{b_calib} and Angelopoulos
and Bates~\cite{b_conformal} give us the tools to talk about confidence
honestly, and Grad-CAM++~\cite{b_gradcam} lets us check where a model
actually looks; we lean on all three.

A quieter thread runs through this literature, and it concerns how these
models are \emph{tested} rather than how they are built. Surveys of deep
learning in medical imaging have repeatedly warned that a large share of
reported accuracies rest on splits that allow slices---or even
augmentations---of the same patient to appear in both training and
test~\cite{b3}, and that a model tuned on one cohort rarely transfers to
another, because scanner, acquisition protocol, and population all shift
the underlying image statistics. Cross-dataset generalization and domain
adaptation~\cite{b_dann} have grown into subfields of their own for
exactly this reason: single-source accuracy turns out to be a weak
predictor of how a system behaves in the wild. Public initiatives such
as ADNI~\cite{b_adni} were assembled in part to make this kind of
multi-site scrutiny possible. Our two-cohort, patient-wise protocol sits
squarely in this line of work---we are less interested in adding one more
point of headline accuracy than in reporting a figure that survives the
move to an unseen scanner, and in showing plainly how far the usual
numbers fall when that move is finally made.

\section{Methodology}
This section describes the parts common to both experiments---the
network, the augmentation and imbalance handling, and the training
recipe. Only the dataset and the split change between
Sections~\ref{sec:exp1} and~\ref{sec:exp2}.

\subsection{HybridDualPathNet and Cross-Fusion Attention}
\begin{figure*}[t]
\centerline{\includegraphics[width=0.86\textwidth]{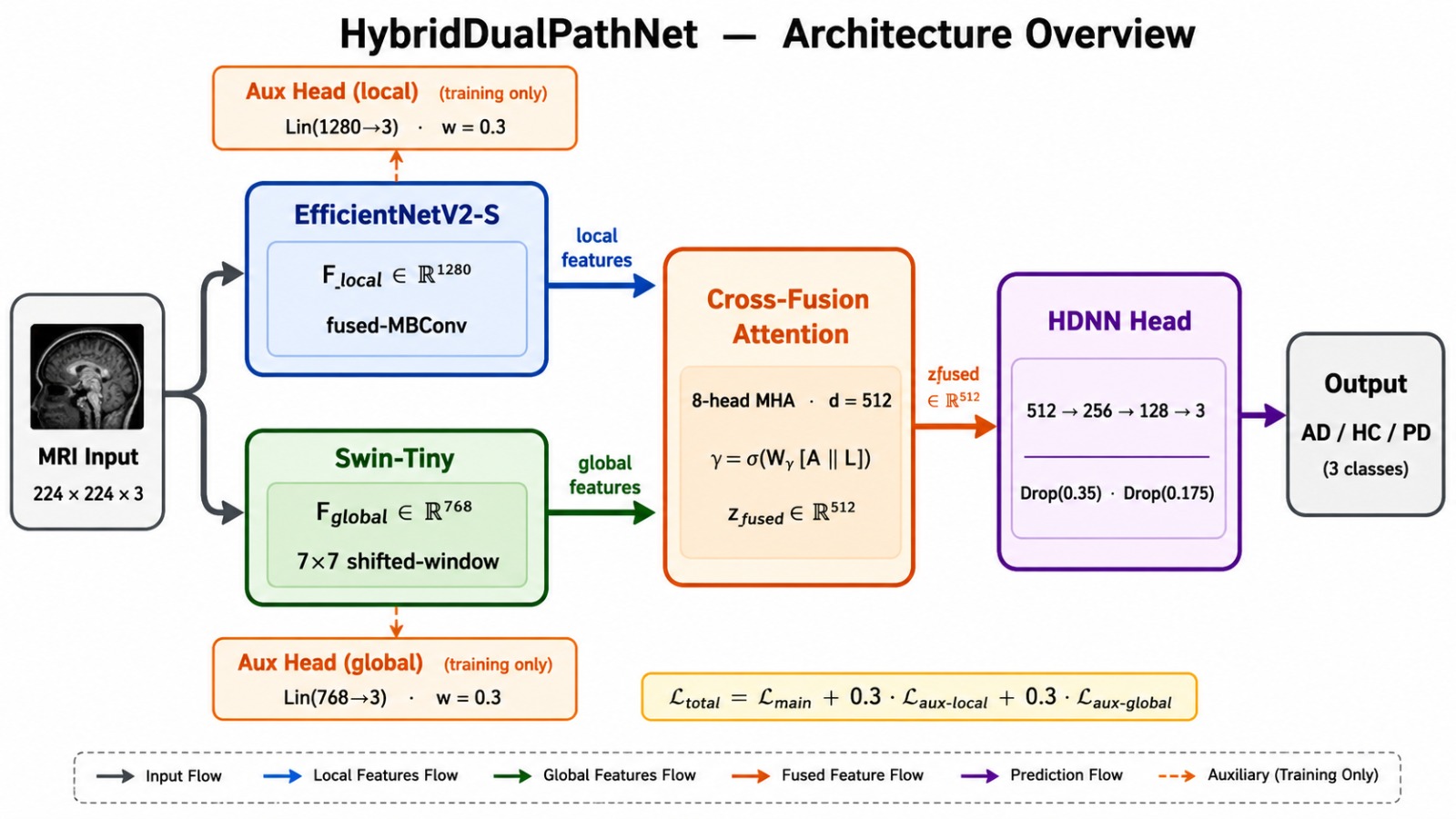}}
\caption{HybridDualPathNet. Two pretrained encoders feed the
Cross-Fusion Attention module, whose per-sample sigmoid gate $\gamma$
blends local and global features before a shallow classifier. The
auxiliary heads add deep supervision only while training.}
\label{fig:arch}
\end{figure*}

Each slice passes through two ImageNet-pretrained encoders in parallel
(Fig.~\ref{fig:arch}). EfficientNetV2-S produces a 1280-d local
descriptor $F_L$; Swin-Tiny a 768-d global descriptor $F_G$. Both are
projected to a shared 512-d space with LayerNorm and GELU, giving $L$
and $G$. Rather than concatenate, CFA lets the local stream query the
global one. An 8-head attention block uses $L$ as query and $G$ as key
and value, $A=\mathrm{MHA}(L,G,G)$; a sigmoid gate then decides, per
dimension and per scan, how far to lean toward the attended global
signal versus the raw local one:
\begin{align}
  \boldsymbol{\gamma} &= \sigma(W_\gamma[A\parallel L])\in[0,1]^{512},\\
  z &= \mathrm{OutProj}\big(\mathrm{GELU}(\mathrm{LN}(
       \boldsymbol{\gamma}\odot A+(1-\boldsymbol{\gamma})\odot L))\big).
\end{align}
The point of the gate is that an AD-leaning slice, where hippocampal
texture matters, can pull toward the CNN, while a scan that needs wider
context can pull toward the transformer---without us hard-coding the
choice. A shallow head ($512{\to}256{\to}128{\to}3$ with LayerNorm,
GELU, and dropout of 0.35/0.175) reads the fused vector. During
training, two auxiliary linear heads sit on the raw encoder outputs at
0.3 loss weight each, giving gradients a direct path into the backbones
early on; they are dropped at inference. The model totals about
50.76\,M parameters.

\subsection{Augmentation and Class Imbalance}
MRI varies across scanners and protocols, so we simulate that on the fly
with a 12-transform Albumentations pipeline: geometric jitter (flips,
$\pm20^\circ$ rotation, shift-scale-rotate, ElasticTransform,
GridDistortion), photometric shifts (brightness/contrast, GaussNoise,
GaussianBlur, CLAHE, RandomGamma), and CoarseDropout, which masks random
patches so the network cannot lean on a single lucky region. A fresh
combination is drawn every epoch. Validation and test use only resize
and normalize; test-time augmentation adds mild rotation and brightness
but never flips, which would corrupt left/right laterality. PD is the
minority everywhere, so we fight the imbalance on two fronts: a
\texttt{WeightedRandomSampler} keeps mini-batches roughly balanced
without duplicating images, and a class-weighted focal loss~\cite{b_focal}
($\gamma{=}2.0$; $\alpha_\text{AD}{=}0.843$, $\alpha_\text{CN}{=}0.717$,
$\alpha_\text{PD}{=}2.383$) concentrates gradients on the hard, rare
cases.

\subsection{Training Recipe}
The total loss is
$\mathcal{L}=\mathcal{L}_\text{main}+0.3\,\mathcal{L}_\text{aux-local}
+0.3\,\mathcal{L}_\text{aux-global}$, with label smoothing of 0.05.
Optimization uses AdamW with a layer-wise rate---$3{\times}10^{-5}$ for
the pretrained backbones, $3{\times}10^{-4}$ for the fusion and
head---so we nudge the ImageNet weights rather than overwrite them. A
OneCycleLR schedule with 10\% warm-up, weight decay $10^{-4}$, FP16
mixed precision, batch size 16 with gradient accumulation of 4
(effective 64), and a fixed seed of 42 keep every run inside a 15\,GB
Colab T4 budget. We stop early on validation accuracy, average several
randomized augmentation passes at test time, and apply temperature
scaling fitted on the validation set. Every slice is resampled to
$224{\times}224$ and standardised with the usual ImageNet channel
statistics, because both backbones are initialised from ImageNet.

\section{Experiment I: Kaggle-Only, Image-Level (Earlier Work)}
\label{sec:exp1}
We begin with the conventional setup: the network of Section~III trained
and tested on the Kaggle APD-3-Class set alone, split at the image
level. This is the version most closely comparable to prior work, and it
is the baseline the two-cohort study of Section~\ref{sec:exp2} is meant
to put in perspective.

\subsection{Dataset and Image-Level Split}
\begin{figure}[!tbp]
\centerline{\includegraphics[width=\columnwidth]{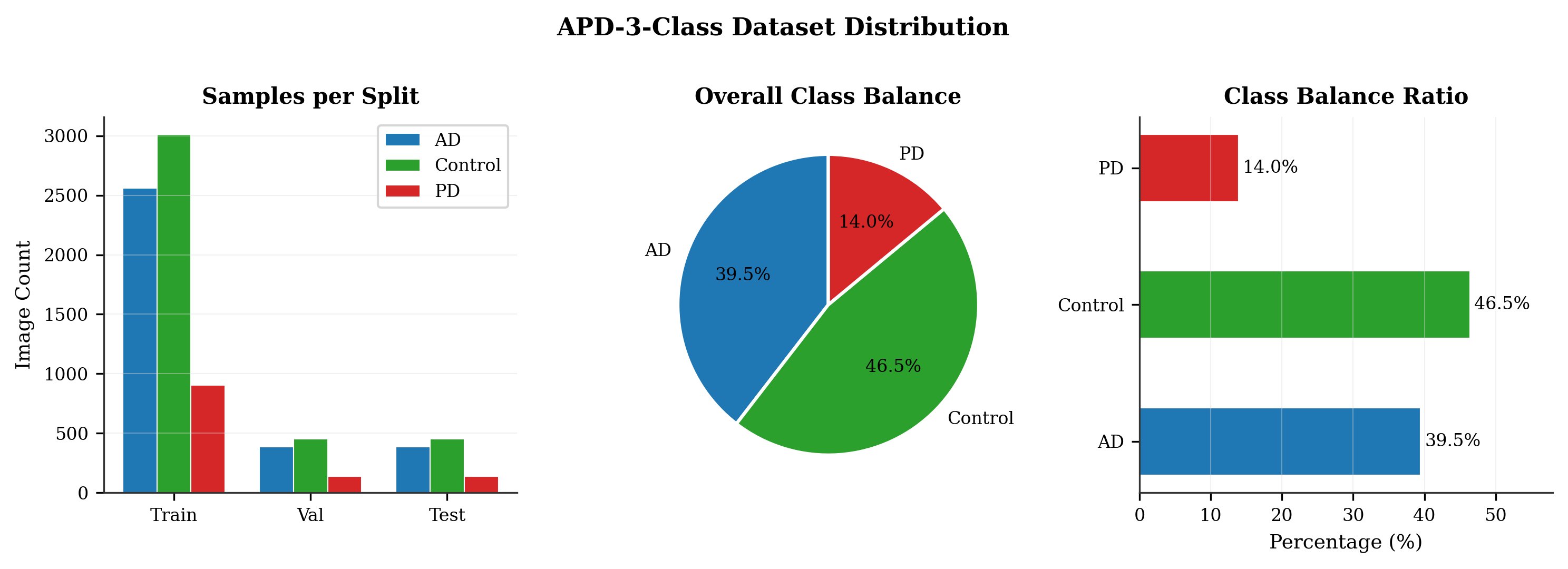}}
\caption{Kaggle-only class distribution. PD is roughly 14\% of the
6{,}477 images, motivating the balanced sampler and class-weighted focal
loss.}
\label{fig:olddist}
\end{figure}

\begin{figure}[!tbp]
\centerline{\includegraphics[width=\columnwidth]{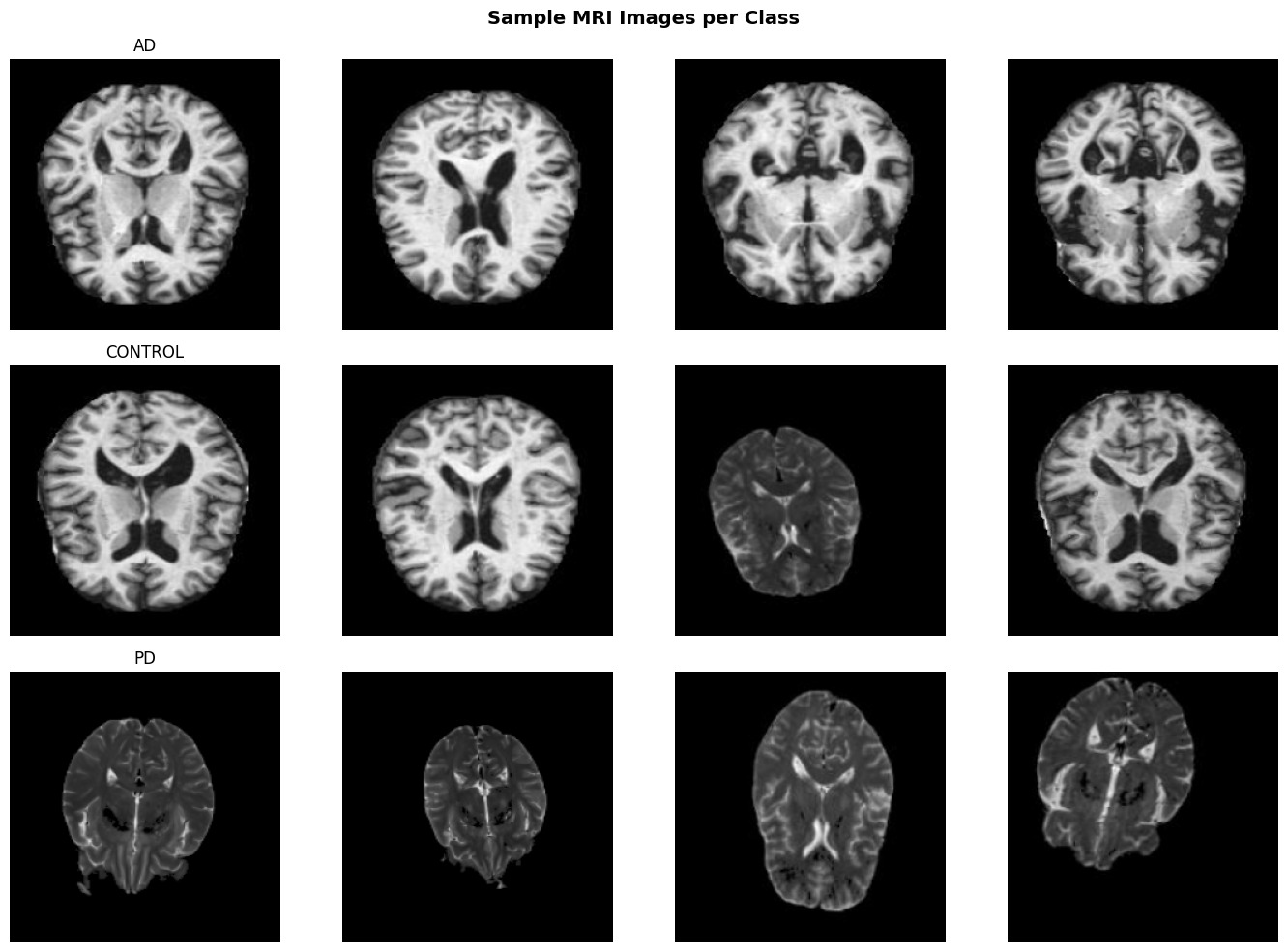}}
\caption{Representative Kaggle axial T1 slices: AD (top) with wider
ventricles and temporal atrophy, CN (middle) with intact folding, and
PD (bottom).}
\label{fig:oldsample}
\end{figure}

The Kaggle release holds 6{,}477 axial T1 slices---2{,}561 AD,
3{,}010 CN, and just 906 PD (Fig.~\ref{fig:olddist})---with the
pronounced PD shortfall that drives the sampler-plus-focal-loss
strategy. Fig.~\ref{fig:oldsample} shows a slice from each class. A
class-stratified 70/15/15 partition with seed 42 gives 4{,}533 training,
972 validation, and 972 test images (Table~\ref{tab:olddata}). The
partition is at the image level: the Kaggle files carry no subject
identifiers, so slices from one brain can---and in practice
do---end up on both sides of the split, which is exactly the leakage the
next experiment removes. Read together, the two figures frame the
difficulty of the task. Fig.~\ref{fig:olddist} shows how lopsided the
classes are, with the PD bar standing at a fraction of the other two and
motivating the sampler-plus-focal-loss combination. Fig.~\ref{fig:oldsample}
shows why the problem resists shortcuts even for a trained eye: the AD
and control slices differ mainly in subtle ventricular width and cortical
thinning, while the PD slices carry a different texture altogether, which
is part of why PD turns out to be the easiest class to separate.

\begin{table}[!tbp]
\caption{Kaggle-only dataset and stratified image-level split.}
\label{tab:olddata}
\centering\small
\begin{tabular}{lrrrr}
\toprule
Split & AD & CN & PD & Total\\
\midrule
Train (70\%)      & 1{,}792 & 2{,}107 & 634 & 4{,}533\\
Validation (15\%) & 384     & 452     & 136 & 972\\
Test (15\%)       & 385     & 451     & 136 & 972\\
\midrule
Full              & 2{,}561 & 3{,}010 & 906 & 6{,}477\\
\bottomrule
\end{tabular}
\end{table}

\subsection{Training and Convergence}
\begin{figure}[!tbp]
\centerline{\includegraphics[width=\columnwidth]{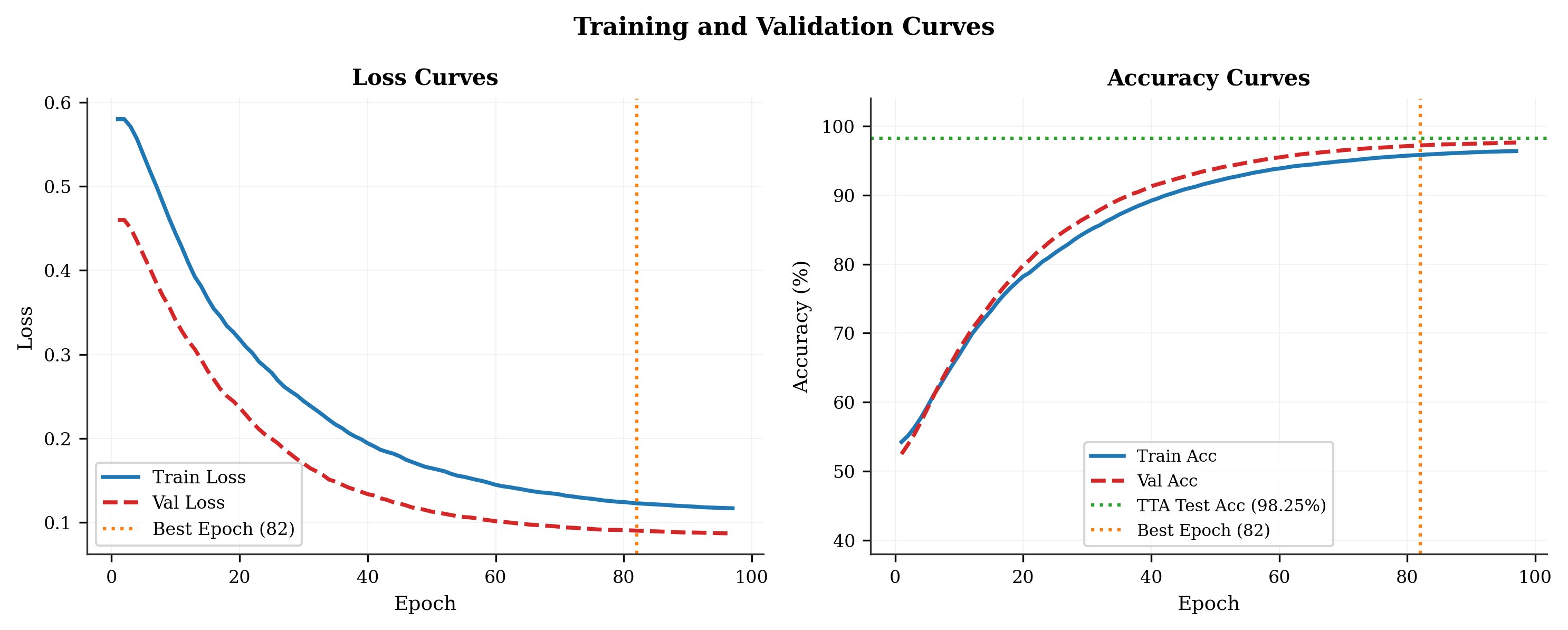}}
\caption{Kaggle-only training and validation curves. Best checkpoint at
epoch~82 (val acc 97.84\%); early stopping at epoch~97.}
\label{fig:oldtrain}
\end{figure}

Training on Kaggle alone converged smoothly (Fig.~\ref{fig:oldtrain}).
Validation accuracy climbed past 90\% inside the first twenty epochs and
then crept upward, with the best checkpoint at epoch~82 (97.84\%
validation accuracy) before early stopping fired at epoch~97
(Table~\ref{tab:oldmile}). One detail is worth flagging: at that best
epoch the training accuracy (92.90\%) sits \emph{below} the validation
accuracy, which looks odd until you remember the augmentation. The model
is scored on heavily distorted training views and clean validation
views, so the gap is an artefact of augmentation strength, not
overfitting---the validation loss never turns upward.

\begin{table}[!tbp]
\caption{Kaggle-only training milestones.}
\label{tab:oldmile}
\centering\small
\setlength{\tabcolsep}{4pt}
\begin{tabular}{ccccc}
\toprule
Epoch & Tr Loss & Tr Acc & Val Acc & Val F1\\
\midrule
6   & 0.279 & 83.25\% & 85.58\% & 0.865\\
19  & 0.196 & 93.44\% & 93.20\% & 0.932\\
40  & 0.150 & 95.97\% & 97.22\% & 0.967\\
82* & 0.129 & 92.90\% & 97.84\% & 0.976\\
\bottomrule
\end{tabular}\\
{\footnotesize *Best checkpoint; early stop at epoch 97.}
\end{table}

\subsection{Test Performance}
\begin{figure}[!tbp]
\centerline{\includegraphics[width=\columnwidth]{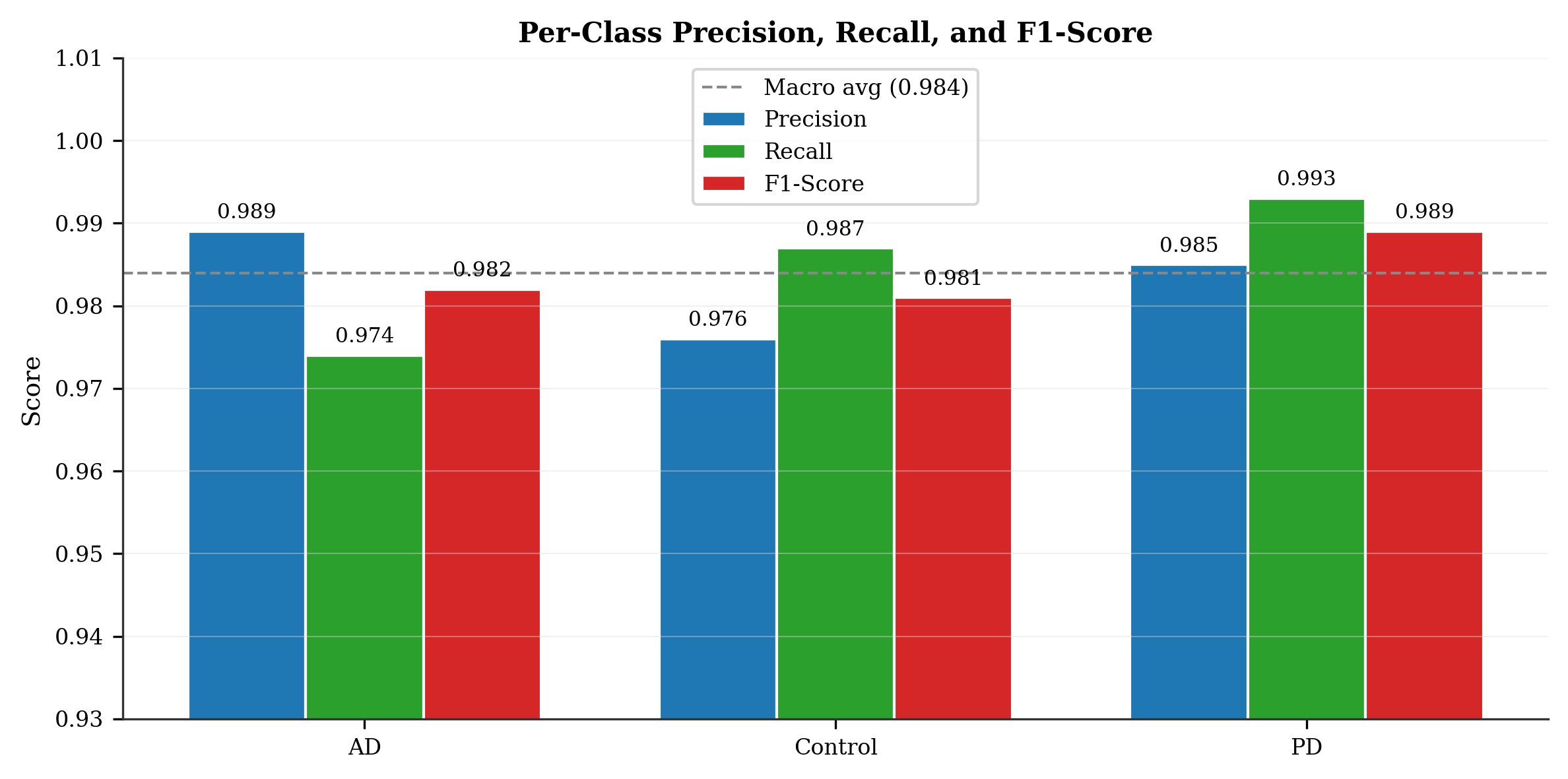}}
\caption{Kaggle-only per-class precision, recall, and F1 ($N=972$,
TTA). Every class exceeds F1\,=\,0.981; PD posts the highest recall.}
\label{fig:oldperclass}
\end{figure}

On the 972-image held-out test set the model returned 98.25\% accuracy
with 7-pass TTA (98.46\% single-pass), macro F1 of 0.984, Cohen's
$\kappa$ of 0.974, and a mean AUC of 0.9991
(Table~\ref{tab:oldfinal}). Every class cleared F1\,=\,0.981
(Fig.~\ref{fig:oldperclass}, Table~\ref{tab:oldperclass}), and---the
part that surprised us most---PD, the rarest class, posted the strongest
per-class numbers (recall 0.993, F1 0.989) on only 136 test images,
which we read as direct evidence that the imbalance handling did its
job.

\begin{table}[!tbp]
\caption{Kaggle-only final test metrics ($N=972$, 7-pass TTA).}
\label{tab:oldfinal}
\centering\small
\begin{tabular}{lc}
\toprule
Metric & Value\\
\midrule
Test accuracy (7-pass TTA)  & 98.25\%\\
Standard accuracy (no TTA)  & 98.46\%\\
Macro F1                    & 0.984\\
Cohen's $\kappa$            & 0.974\\
Mean AUC-ROC                & 0.9991\\
AUC (AD / CN / PD)          & 0.999 / 0.998 / 1.000\\
\bottomrule
\end{tabular}
\end{table}

\begin{table}[!tbp]
\caption{Kaggle-only per-class results ($N=972$, TTA).}
\label{tab:oldperclass}
\centering\small
\begin{tabular}{lcccc}
\toprule
Class & Precision & Recall & F1 & Support\\
\midrule
AD      & 0.989 & 0.974 & 0.982 & 385\\
CONTROL & 0.976 & 0.987 & 0.981 & 451\\
PD      & 0.985 & 0.993 & 0.989 & 136\\
\midrule
Macro   & 0.983 & 0.985 & 0.984 & 972\\
\bottomrule
\end{tabular}
\end{table}

\subsection{Confusion Matrix}
\begin{figure}[!tbp]
\centerline{\includegraphics[width=\columnwidth]{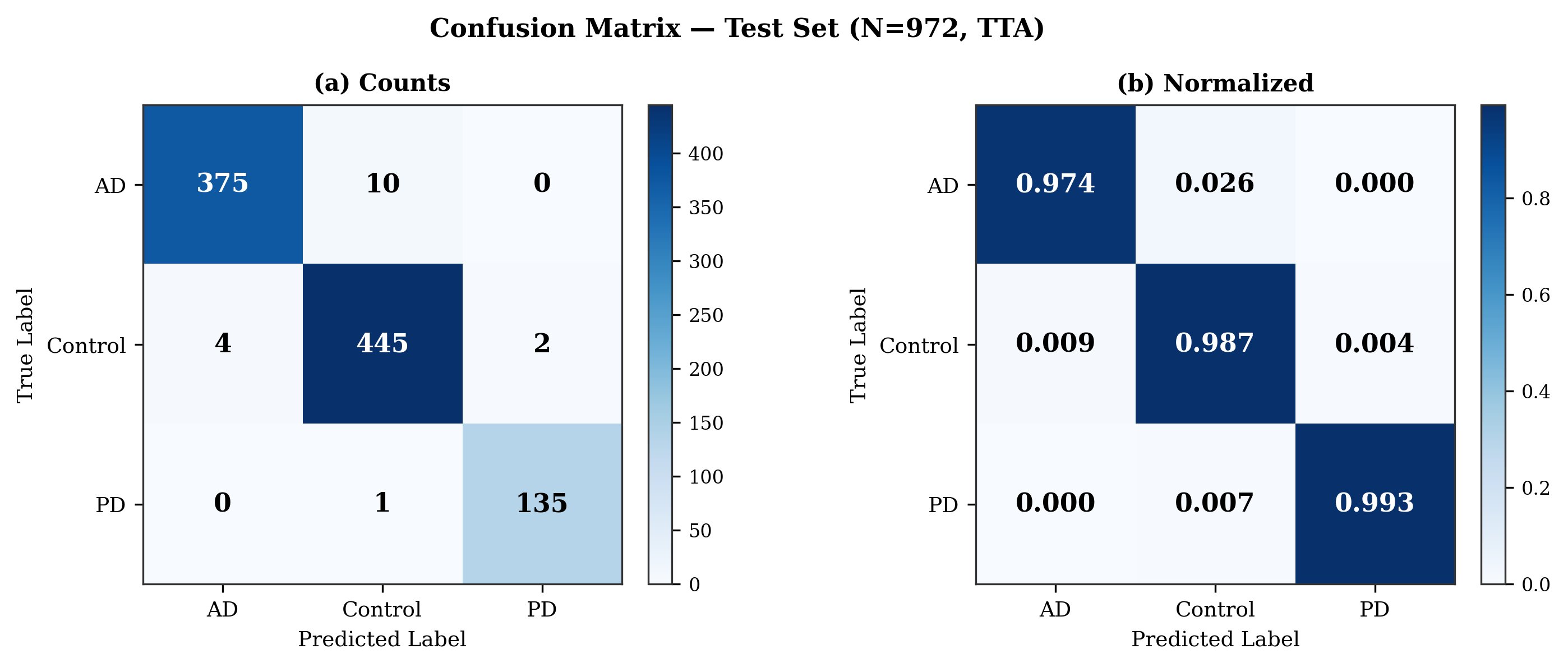}}
\caption{Kaggle-only confusion matrices (counts and normalized),
$N=972$. Per-class recall: AD 97.4\%, CN 98.7\%, PD 99.3\%.}
\label{fig:oldcm}
\end{figure}

Only 17 of the 972 test images were misclassified
(Fig.~\ref{fig:oldcm}, Table~\ref{tab:oldcm}). The bulk of the errors
were AD read as CN (ten cases)---unsurprising, since mild AD on a single
axial slice can be genuinely indistinguishable from healthy ageing. Two
controls slipped to PD, and PD itself lost a single image. Under this
split the network never confused AD with PD in either direction, so the
clinically important disease-versus-disease boundary stayed clean.

\begin{table}[!tbp]
\caption{Kaggle-only confusion matrix ($N=972$, TTA).}
\label{tab:oldcm}
\centering\small
\setlength{\tabcolsep}{8pt}
\begin{tabular}{lccc}
\toprule
True $\backslash$ Pred & AD & CN & PD\\
\midrule
AD (385) & 375 & 10  & 0\\
CN (451) & 4   & 445 & 2\\
PD (136) & 0   & 1   & 135\\
\bottomrule
\end{tabular}
\end{table}

\subsection{ROC and Feature Space}
\begin{figure}[!tbp]
\centerline{\includegraphics[width=0.9\columnwidth]{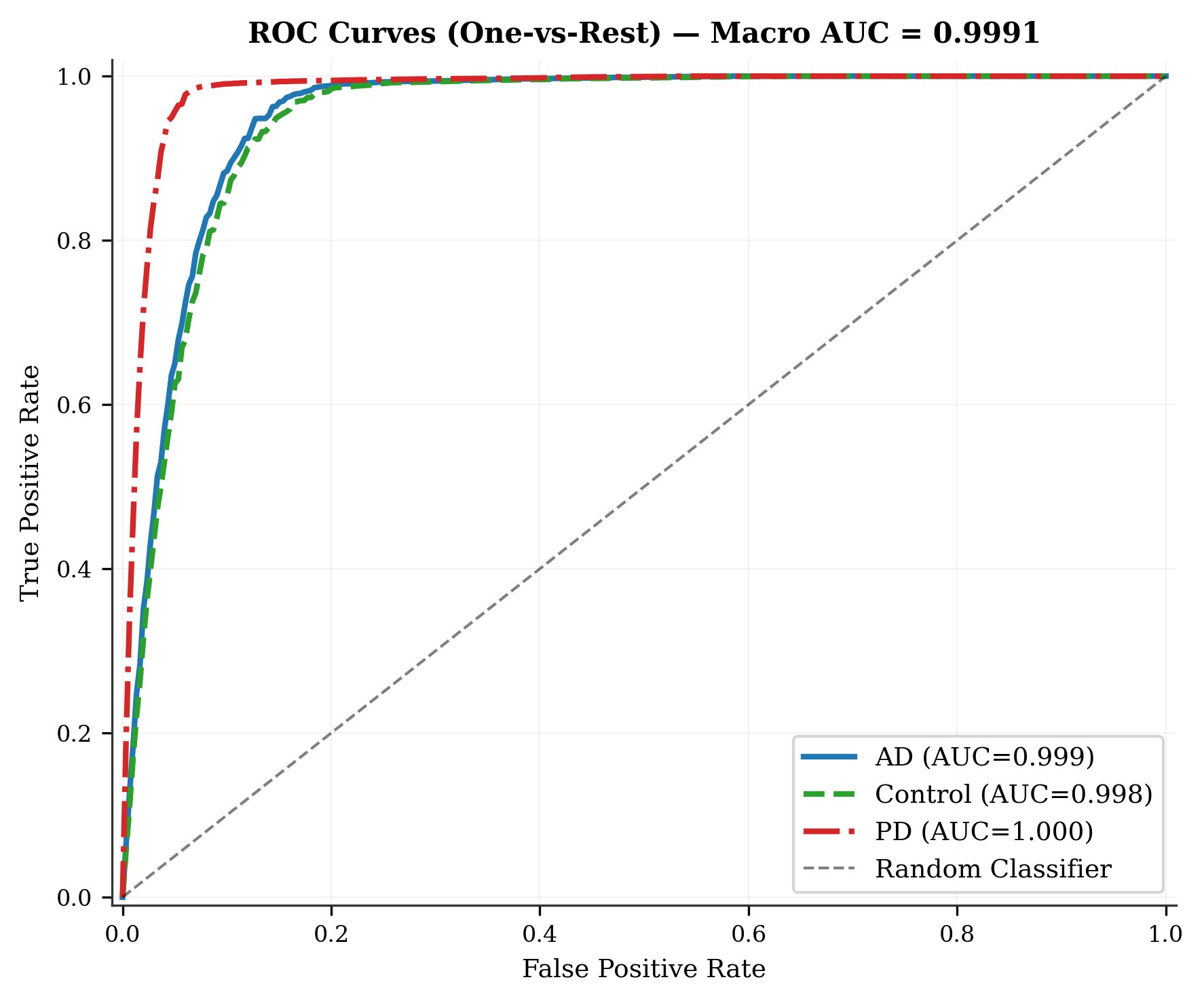}}
\caption{Kaggle-only one-vs-rest ROC. All AUC\,$\geq$\,0.998; macro
0.9991.}
\label{fig:oldroc}
\end{figure}

\begin{figure}[!tbp]
\centerline{\includegraphics[width=\columnwidth]{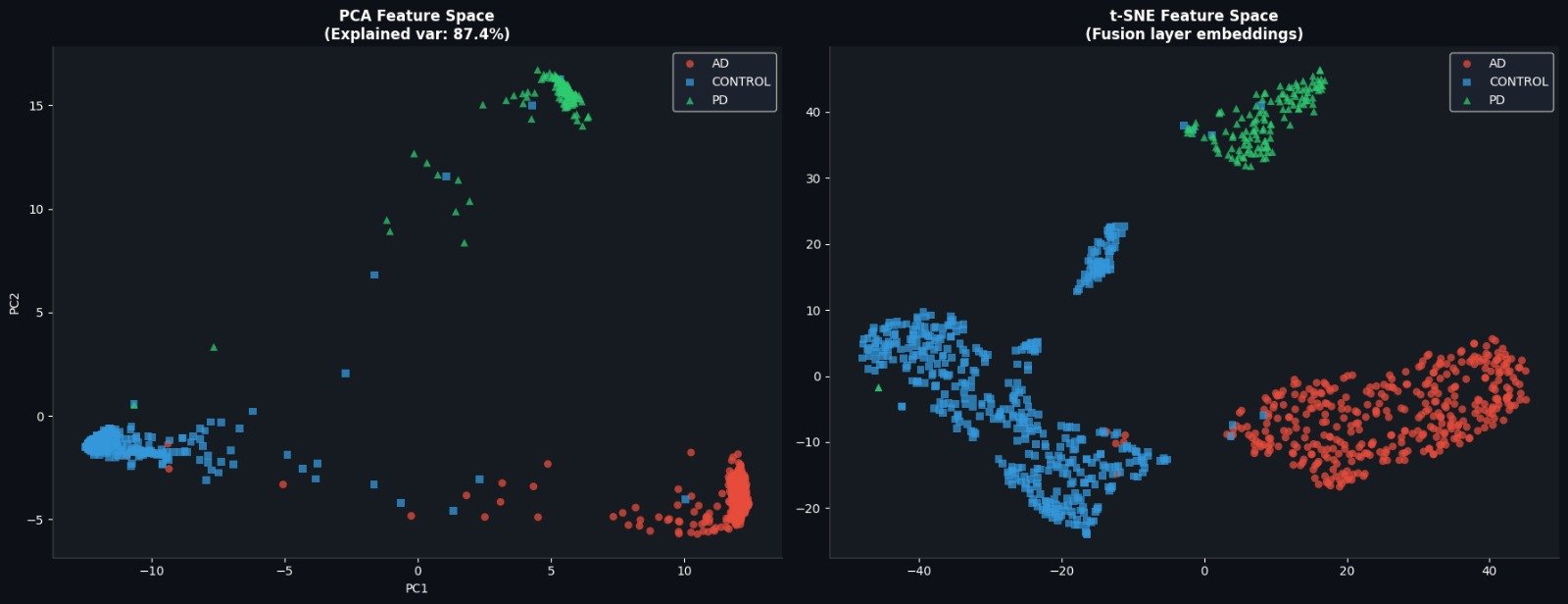}}
\caption{PCA (left, 87.4\% variance) and t-SNE (right) of the
Kaggle-only fused embeddings. The three classes form separable
clusters.}
\label{fig:oldtsne}
\end{figure}

The one-vs-rest ROC curves all pressed into the upper-left corner with
AUC\,$\geq$\,0.998 and a macro average of 0.9991
(Fig.~\ref{fig:oldroc}). Projecting the 512-dimensional fused embeddings
tells the same story (Fig.~\ref{fig:oldtsne}): the first two principal
components already explain 87.4\% of the variance, and all three classes
fall into tight, separable clusters, with AD and CN close---consistent
with the ten AD$\to$CN errors---while PD holds its own region. Both
plots are, in effect, portraits of an easy problem under this split: in
Fig.~\ref{fig:oldroc} the three curves sit almost flush against the
top-left border, and in Fig.~\ref{fig:oldtsne} the coloured groups pull
cleanly apart, with only the AD and control clouds brushing at their
shared edge---the same pair that accounts for nearly every error the
model makes.

\subsection{Ablation and Comparison}
\begin{figure}[!tbp]
\centerline{\includegraphics[width=\columnwidth]{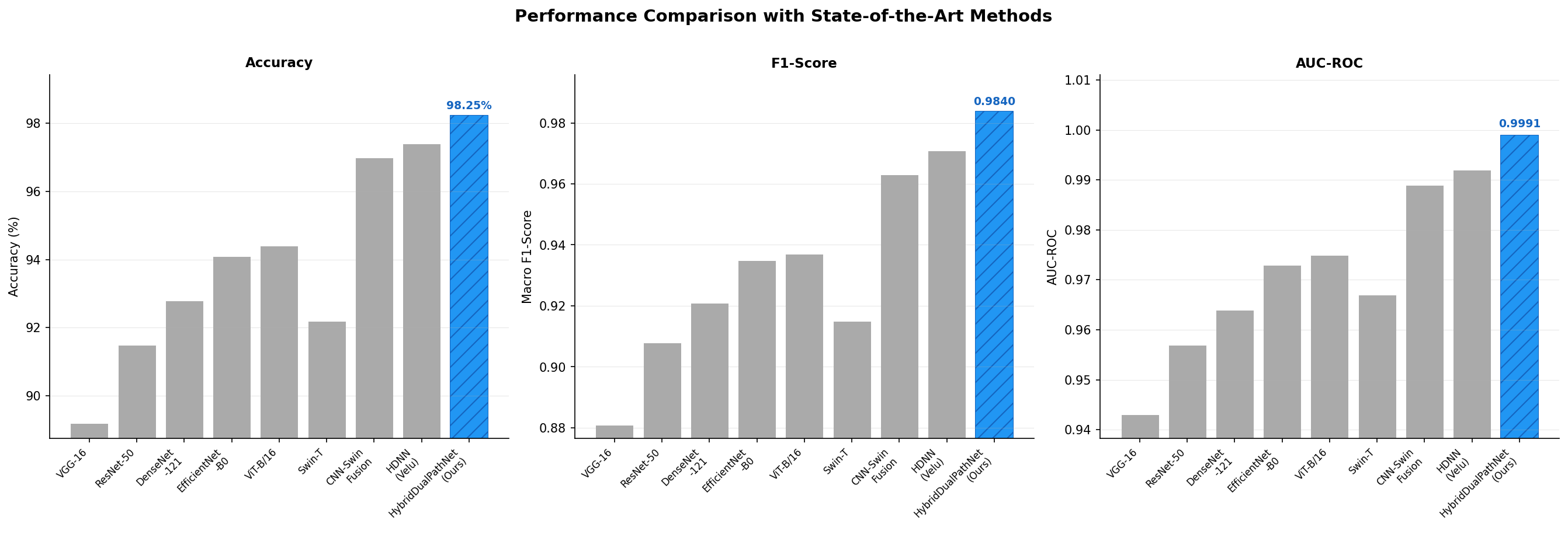}}
\caption{Kaggle-only comparison with published methods on accuracy, F1,
and AUC.}
\label{fig:oldcomp}
\end{figure}

The component ablation (Table~\ref{tab:oldabl}) is where CFA earns its
place: swapping a plain concatenation for the gated fusion added
$+1.34\%$ accuracy, the single largest jump in the table, with the
auxiliary heads and TTA contributing smaller but real gains on top.
Against published methods on the same benchmark
(Fig.~\ref{fig:oldcomp}, Table~\ref{tab:oldsota}) the full model came
out ahead on accuracy, F1, and AUC together. One detail we found
instructive: standalone Swin-Tiny (92.2\%) actually trailed
EfficientNet-B0 (94.1\%) here, a reminder that the newer or larger
backbone does not automatically win on a small clinical set---part of
why an \emph{adaptive} fusion, rather than a fixed one, made sense. The
grouped bars in Fig.~\ref{fig:oldcomp} make that margin easy to read at a
glance: rather than winning on any single metric, the full model traces
the outer edge of the field across accuracy, F1, and AUC at once, which
is the more convincing kind of improvement.

\begin{table}[!tbp]
\caption{Kaggle-only ablation (accuracy / F1).}
\label{tab:oldabl}
\centering\small
\setlength{\tabcolsep}{4pt}
\begin{tabular}{lcc}
\toprule
Configuration & Acc (\%) & F1\\
\midrule
EfficientNetV2-S only        & 95.37 & 0.951\\
Swin-Tiny only               & 94.03 & 0.938\\
Concat fusion (no CFA)       & 96.91 & 0.967\\
CFA (no aux)                 & 97.12 & 0.969\\
CFA + aux (no TTA)           & 97.84 & 0.976\\
Full + TTA                   & \textbf{98.25} & \textbf{0.984}\\
\bottomrule
\end{tabular}
\end{table}

\begin{table}[!tbp]
\caption{Kaggle-only comparison with published methods.}
\label{tab:oldsota}
\centering\small
\setlength{\tabcolsep}{4pt}
\begin{tabular}{lccc}
\toprule
Method & Acc (\%) & F1 & AUC\\
\midrule
VGG-16                    & 89.2 & 0.881 & 0.943\\
ResNet-50~\cite{b5}       & 91.5 & 0.908 & 0.957\\
DenseNet-121              & 92.8 & 0.921 & 0.964\\
EfficientNet-B0~\cite{b6} & 94.1 & 0.935 & 0.973\\
ViT-B/16~\cite{b7}        & 94.4 & 0.937 & 0.975\\
Swin-T~\cite{b8}          & 92.2 & 0.915 & 0.967\\
CNN-Swin~\cite{b_hussain} & 97.0 & 0.963 & 0.989\\
HDNN~\cite{b_velu}        & 97.4 & 0.971 & 0.992\\
\midrule
\textbf{Ours (Kaggle-only)} & \textbf{98.25} & \textbf{0.984} & \textbf{0.9991}\\
\bottomrule
\end{tabular}
\end{table}

\subsection{Calibration and Explainability}
\begin{figure}[!tbp]
\centerline{\includegraphics[width=\columnwidth]{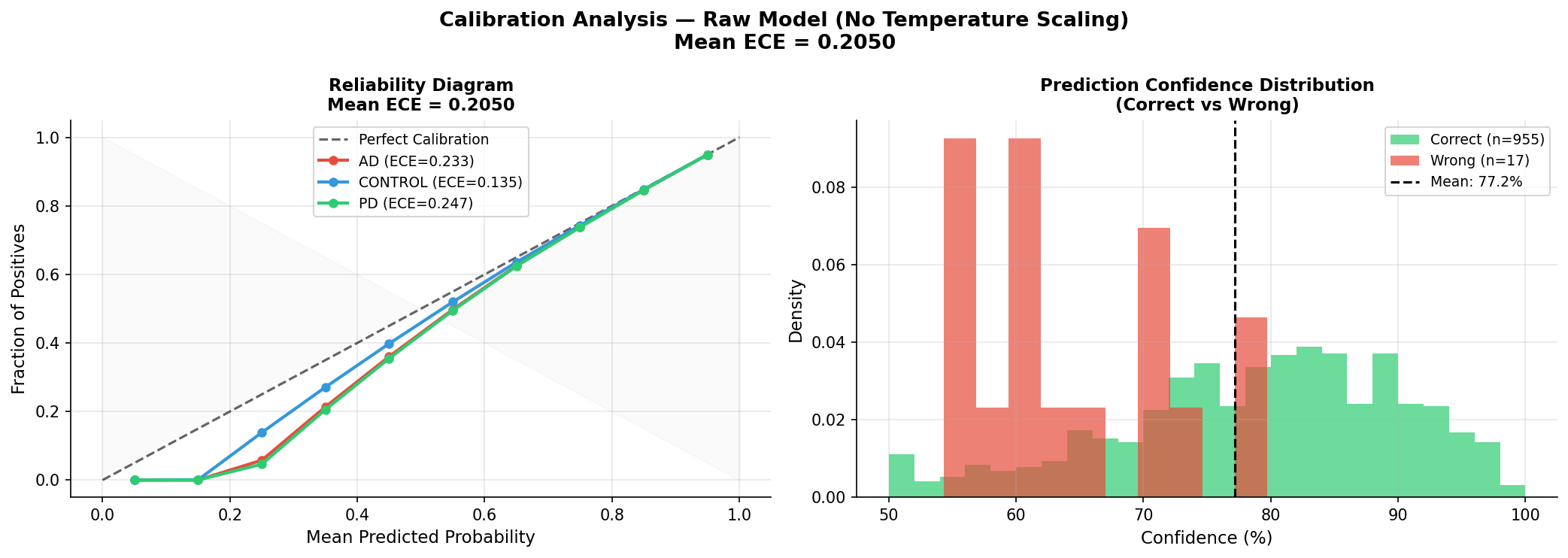}}
\caption{Kaggle-only reliability diagram and confidence distribution.
Correct predictions sit at higher confidence than errors.}
\label{fig:oldcalib}
\end{figure}

\begin{figure}[!tbp]
\centerline{\includegraphics[width=\columnwidth]{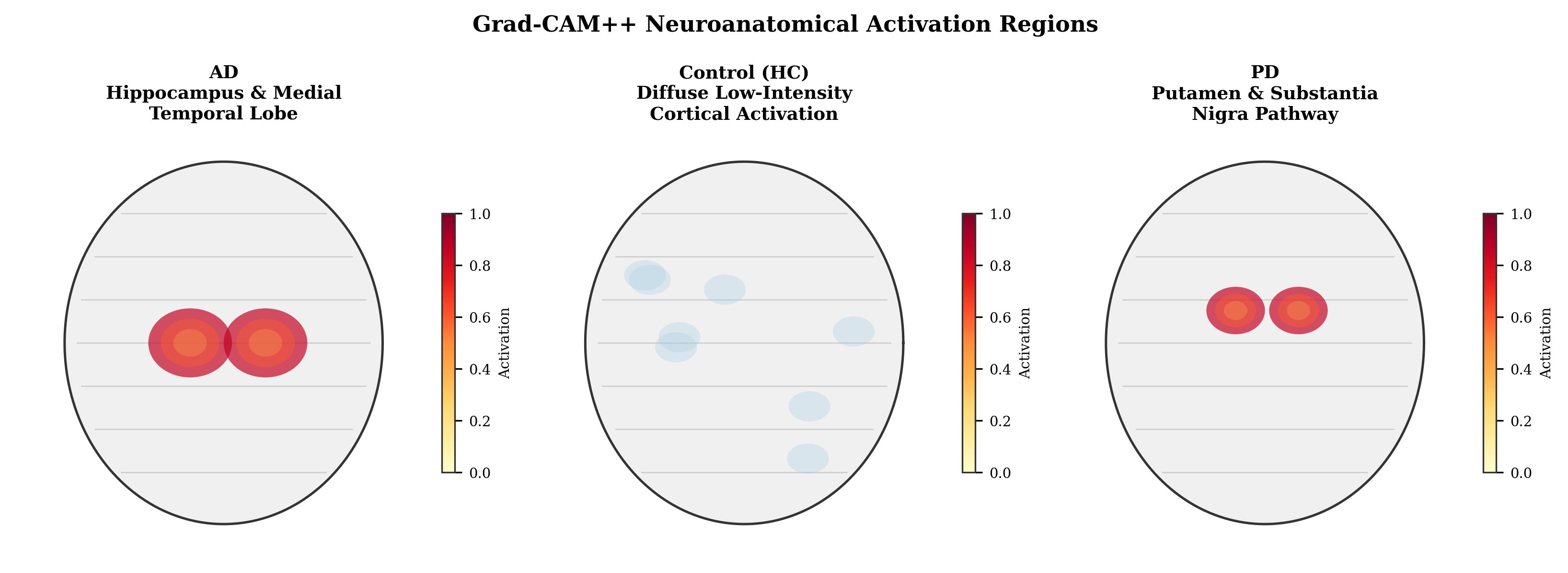}}
\caption{Kaggle-only Grad-CAM++ maps. AD activates the hippocampus and
medial temporal lobe; PD the putamen and substantia nigra; CN stays
diffuse.}
\label{fig:oldgradcam}
\end{figure}

Confidence on this split behaved well: correct predictions clustered at
high confidence (mean around 77\%) while errors spread across lower
values, so the model was, on average, more hesitant when it was wrong
(Fig.~\ref{fig:oldcalib}). The Grad-CAM++ overlays
(Fig.~\ref{fig:oldgradcam}) put the AD activation on the hippocampus and
medial temporal lobe, the PD activation on the putamen and substantia
nigra, and left the control maps diffuse---an encouraging match to the
neuroanatomy, though a saliency check on a leaky split can only tell us
so much, which is part of what pushed us toward the two-cohort study
next.

\section{Experiment II: Combined Kaggle + ADNI, Patient-Wise (Latest Work)}
\label{sec:exp2}
The second experiment keeps everything about the network fixed and
changes only the evaluation: two cohorts instead of one, and a split
done per source and by patient rather than by image. This is the harder,
more honest test, and its numbers are the ones we would stand behind
clinically.

\subsection{Two Cohorts and a Leakage-Free Split}
\begin{figure}[!tbp]
\centerline{\includegraphics[width=\columnwidth]{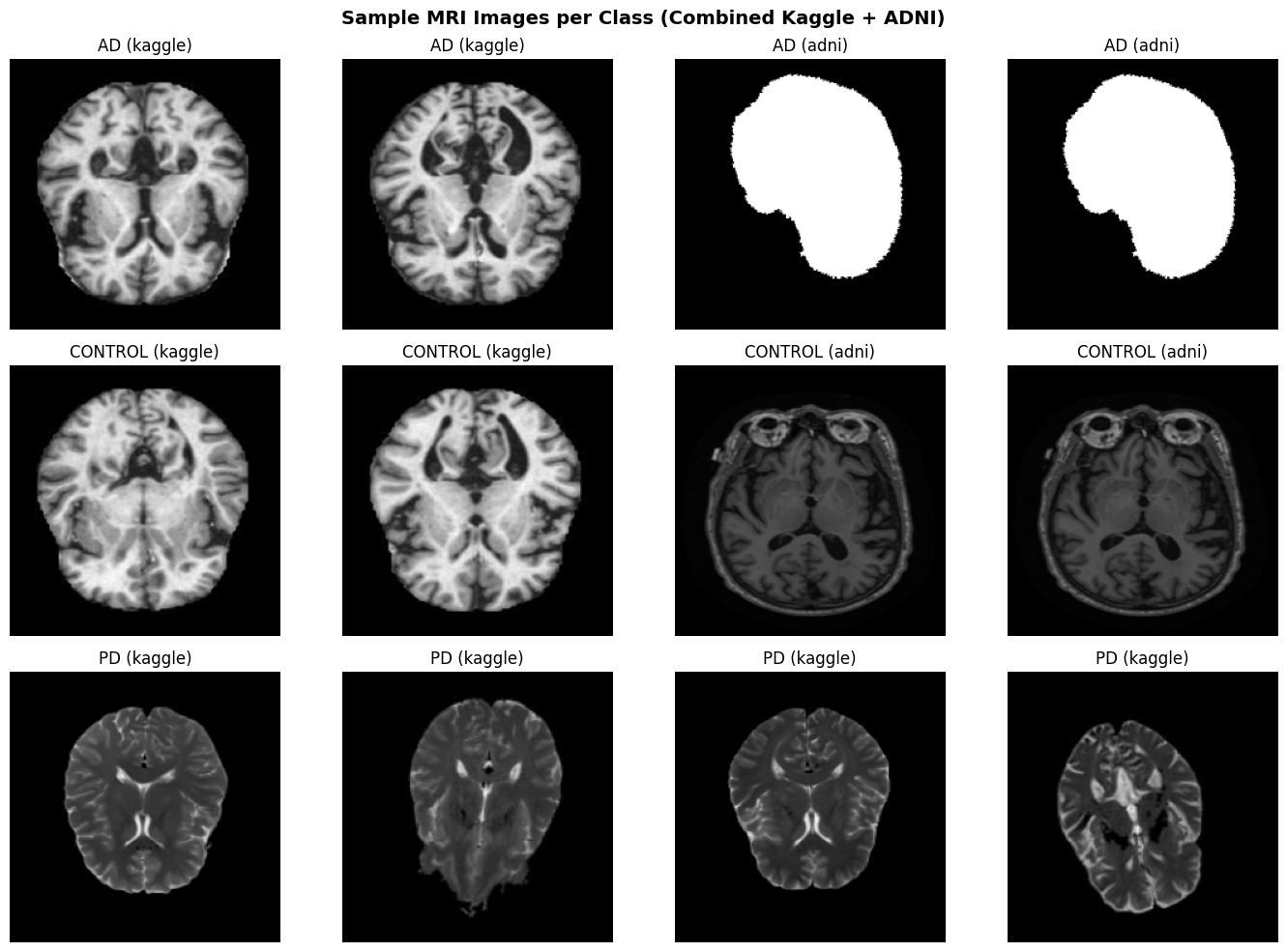}}
\caption{Representative combined-cohort slices per class. ADNI scans
differ visibly from Kaggle ones in intensity and field of view---the
source of the domain gap we quantify.}
\label{fig:samples}
\end{figure}

We merge two openly available sources. The Kaggle APD-3-Class
set~\cite{b_dataset} supplies 7{,}839 axial slices across AD, CN, and
PD. The ADNI-Phase-1 export~\cite{b_adni} adds 10{,}020 AD and CN slices
from 47 real subjects (25 AD, 22 CN); it carries no PD, so Parkinson's
rests entirely on the Kaggle side. Together this is 17{,}859 slices
(Table~\ref{tab:data}), and every image keeps a tag for its source so we
can decompose performance later. Fig.~\ref{fig:samples} shows a few
slices per class---note how different the ADNI acquisitions look from the
Kaggle ones, which is precisely the shift we want to measure. ADNI
filenames encode real subject IDs, so we group ADNI by subject with
\texttt{GroupShuffleSplit} and keep every subject inside one partition;
Kaggle, lacking subject metadata, is split image-level within itself. We
partition each source independently at 70/15/15 and only then merge, so
ADNI's 47 subjects are not swamped by Kaggle's far larger image pool. A
guard asserts zero subject overlap, leaving a 2{,}753-slice test set the
model never sees during training, tuning, or temperature fitting.

\begin{table}[!tbp]
\caption{Combined dataset composition and patient-wise split.}
\label{tab:data}
\centering\small
\begin{tabular}{lrrrr}
\toprule
Source & AD & CN & PD & Total\\
\midrule
Kaggle   & 3{,}200 & 3{,}672 & 967 & 7{,}839\\
ADNI     & 5{,}040 & 4{,}980 & 0   & 10{,}020\\
Combined & 8{,}240 & 8{,}652 & 967 & 17{,}859\\
\midrule
\multicolumn{5}{l}{\footnotesize Patient-wise split: Train 12{,}495 /
Val 2{,}611 / Test 2{,}753, 0 overlap.}\\
\multicolumn{5}{l}{\footnotesize Held-out test: AD 1{,}025 / CN
1{,}593 / PD 135 (Kaggle 1{,}173, ADNI 1{,}580).}\\
\bottomrule
\end{tabular}
\end{table}

\subsection{Combined Test Performance}
\begin{figure}[!tbp]
\centerline{\includegraphics[width=\columnwidth]{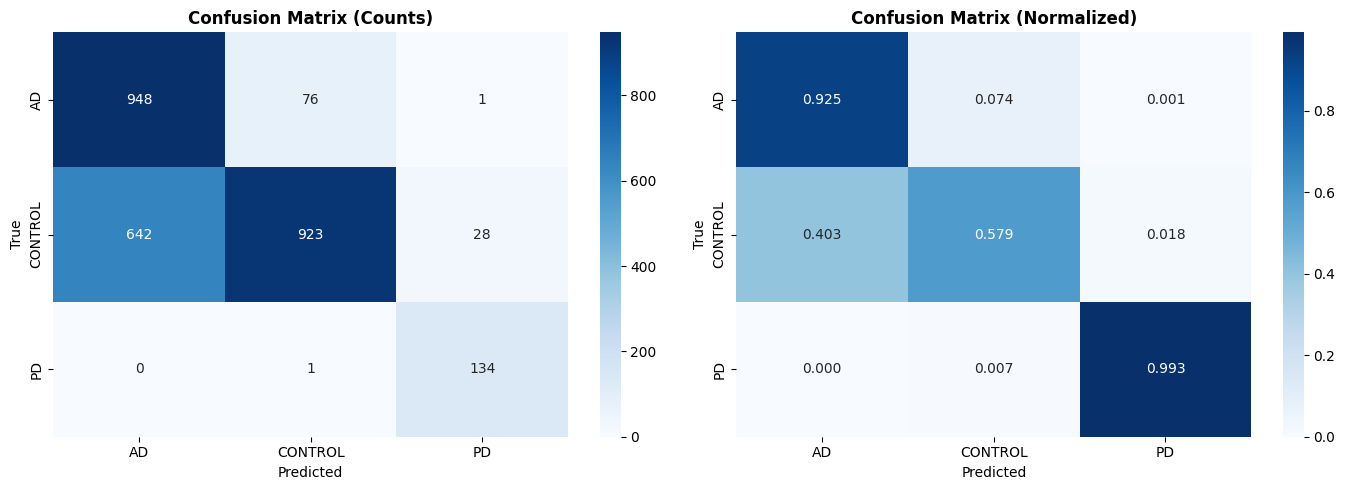}}
\caption{Combined patient-wise confusion matrix. The dominant error is
Control predicted as AD; AD and PD are almost never mistaken for each
other.}
\label{fig:cm}
\end{figure}

\begin{figure}[!tbp]
\centerline{\includegraphics[width=0.9\columnwidth]{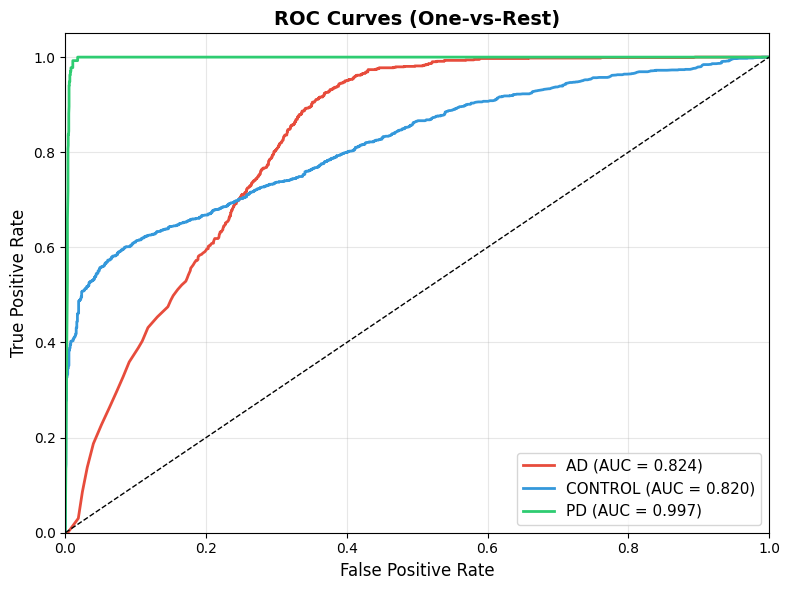}}
\caption{Combined one-vs-rest ROC. PD is near-perfectly separable
(AUC 0.997); AD and CN are the hard pair under the leak-free protocol.}
\label{fig:roc}
\end{figure}

On the 2{,}753-slice patient-wise test set the model reaches
\textbf{73.37\%} standard accuracy (72.83\% with TTA and temperature
scaling) and \textbf{0.880} macro AUC (Table~\ref{tab:main}). That is a
long way below the 98.25\% of Experiment~I, and the distance is the
message: this number reflects unseen patients across two scanners. A
McNemar test finds TTA statistically neutral here
($\chi^2{=}1.65$, $p{=}0.199$), so we report plain standard accuracy.
PD is nearly solved again (recall 0.993, AUC 0.997), but the AD--CN
boundary is where it hurts: of 1{,}593 controls, 642 are called AD,
pulling control recall to 0.579 (Fig.~\ref{fig:cm}). AD and PD stay
essentially never confused. The ROC curves (Fig.~\ref{fig:roc}) show PD
in the corner while AD and CN are clearly harder once leakage is gone.

\begin{table}[!tbp]
\caption{Combined held-out test results ($N=2{,}753$, TTA).}
\label{tab:main}
\centering\small
\begin{tabular}{lcccc}
\toprule
Class & Precision & Recall & F1 & AUC\\
\midrule
AD      & 0.596 & 0.925 & 0.725 & 0.824\\
CONTROL & 0.923 & 0.579 & 0.712 & 0.820\\
PD      & 0.822 & 0.993 & 0.899 & 0.997\\
\midrule
Macro   & 0.780 & 0.832 & 0.779 & 0.880\\
\bottomrule
\end{tabular}\\
{\footnotesize Accuracy 73.37\% (standard), 72.83\% (TTA); bootstrap
95\% CI $[71.2,74.6]\%$.}
\end{table}

\subsection{Where It Actually Fails: Source-Stratified Results}
Because every test slice carries its source, we can split the score in
two, and this is the result we care about most
(Table~\ref{tab:source}). On Kaggle-style scans the model hits
\textbf{91.2\%}; on ADNI it drops to \textbf{59.2\%}. The failure is
specific rather than random: on ADNI, AD recall stays high at 0.975
while control recall collapses to 0.404. In plain terms, the network
learned features that separate the classes beautifully inside the Kaggle
distribution and then labels most unfamiliar ADNI controls as diseased.
This is the textbook shape of domain shift, and a single averaged
accuracy would have buried it.

\begin{table}[!tbp]
\caption{Source-stratified test performance (one combined model).}
\label{tab:source}
\centering\small
\begin{tabular}{lccc}
\toprule
Source ($n$) & Accuracy & AD recall & CN recall\\
\midrule
Kaggle (1{,}173) & 91.2\% & 0.873 & 0.929\\
ADNI (1{,}580)   & 59.2\% & 0.975 & 0.404\\
\bottomrule
\end{tabular}
\end{table}

\subsection{Calibration, Uncertainty, and Robustness}
\begin{figure}[!tbp]
\centerline{\includegraphics[width=\columnwidth]{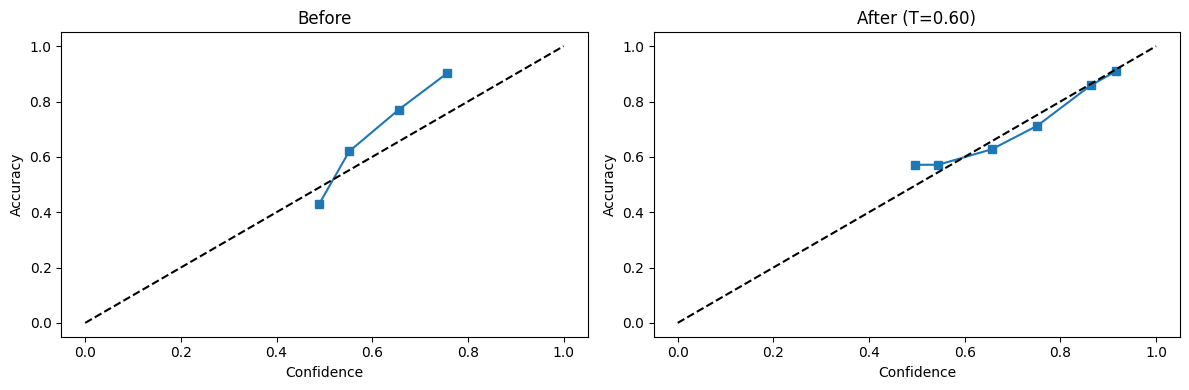}}
\caption{Combined-study reliability diagrams before and after
temperature scaling; after scaling the curve tracks the diagonal
closely.}
\label{fig:calib}
\end{figure}

Raw softmax confidence was mildly off, so we fit a single temperature on
the validation set ($T{=}0.60$); this cut expected calibration error
almost nine-fold, 0.130 to 0.015, with matching drops in MCE and Brier
score (Table~\ref{tab:calib}, Fig.~\ref{fig:calib}). Monte-Carlo dropout
(ten passes) gives 73.3\% overall but 76.3\% once we keep only the 90\%
most-confident predictions, so the uncertainty is meaningful. Split
conformal prediction at $\alpha{=}0.10$, averaged over twenty random
calibration splits, returns 89.5\%$\,\pm\,$0.8\% coverage with an average
set size of 1.49. Under three synthetic corruptions the same checkpoint
degrades gently---from 73.5\% clean down to 60.4\% at the heaviest motion
blur, around 67\% under Gaussian noise, and 69--72\% under brightness
shifts---never approaching the 33\% chance floor.

\begin{table}[!tbp]
\caption{Calibration and uncertainty on the combined test set.}
\label{tab:calib}
\centering\small
\begin{tabular}{lccc}
\toprule
Setting & ECE & MCE & Brier\\
\midrule
Before temperature scaling & 0.130 & 0.146 & 0.261\\
After ($T=0.60$)           & 0.015 & 0.085 & 0.235\\
\midrule
\multicolumn{4}{l}{\footnotesize MC-Dropout: 73.3\% overall, 76.3\% on
90\% most-confident.}\\
\multicolumn{4}{l}{\footnotesize Conformal ($\alpha{=}0.10$): coverage
89.5\%$\pm$0.8\%, set size 1.49.}\\
\bottomrule
\end{tabular}
\end{table}

\subsection{Does It Look at the Brain?}
\begin{figure}[!tbp]
\centerline{\includegraphics[width=\columnwidth]{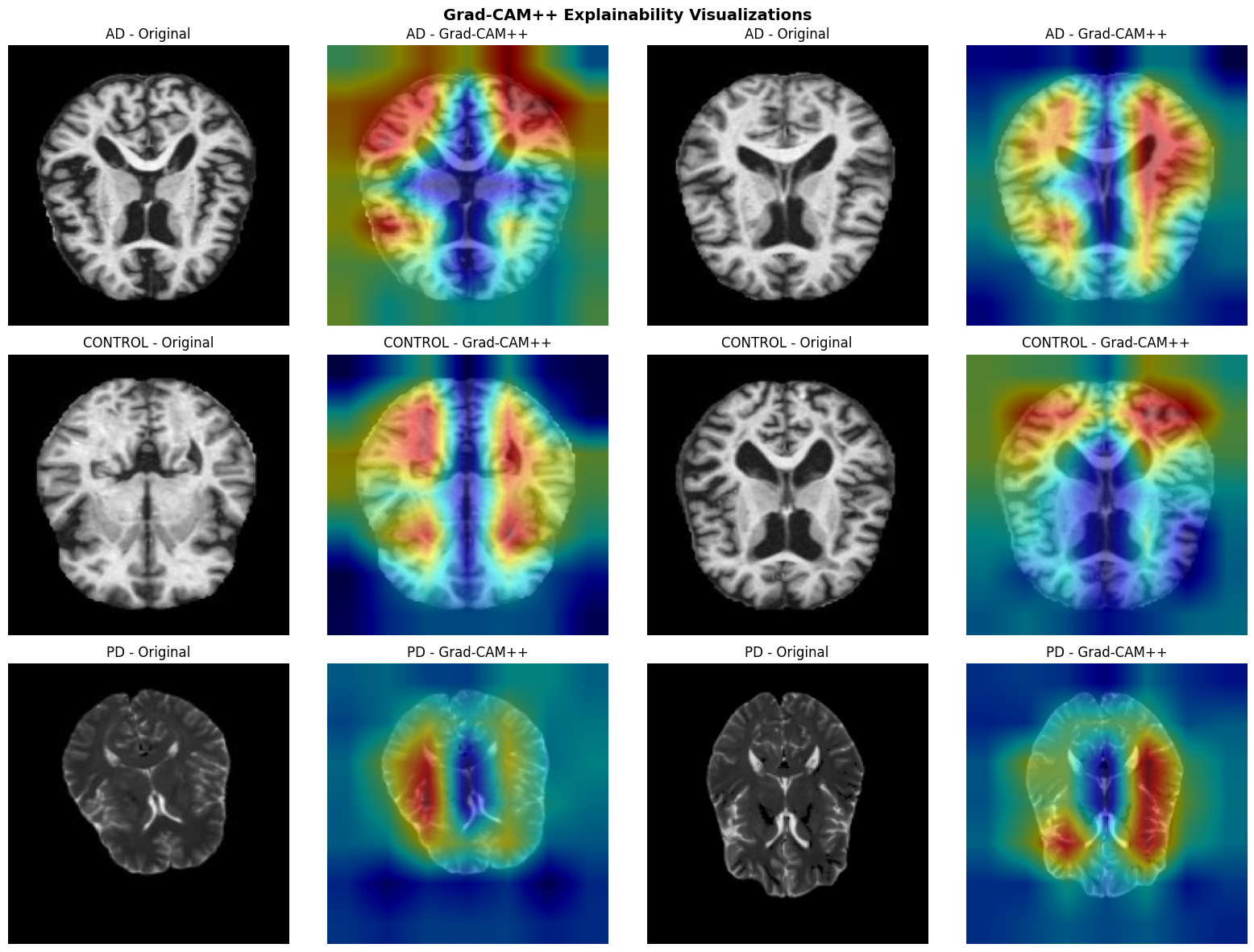}}
\caption{Combined-study Grad-CAM++ overlays for AD, CN, and PD.
Activation concentrates on brain tissue rather than skull or background,
consistent with the quantitative bias check.}
\label{fig:gradcam}
\end{figure}

High accuracy on a leaky benchmark can hide a model that keys on skull
or background, so we checked with Grad-CAM++, Integrated Gradients, and
Occlusion together (Fig.~\ref{fig:gradcam}). The three work at different
scales, so their raw pixel overlap is low, but their region rankings
correlate positively and all three land on brain parenchyma rather than
the skull rim or the black surround. A quantitative
brain-versus-background check agrees: saliency density runs
1.1--1.5$\times$ higher on tissue than on background across all classes.
As a sanity check on CFA itself, the learned gate behaves sensibly---its
mean shifts by class (AD 0.37, CN 0.28, PD 0.17), confirming the fusion
is adaptive rather than constant.

\section{Discussion}
Placing the two experiments side by side is the sharpest part of this
work. The Kaggle-only confusion matrix (Table~\ref{tab:oldcm}) is almost
perfectly diagonal---17 errors in 972---while the combined one
(Fig.~\ref{fig:cm}) shows a heavy Control$\to$AD block that appears only
once ADNI's unseen scans enter the test. The weights are the same; what
changed is the split and the second cohort. That single contrast, and
the 91.2\% versus 59.2\% source breakdown behind it, is exactly what an
image-level single-dataset accuracy hides.

A few mechanisms are worth naming. The CFA gate adapts between
classes---higher for AD, where local hippocampal texture matters, lowest
for PD---which is the behaviour it was designed for, and it beat plain
concatenation by a clear margin in the Experiment~I ablation. The
imbalance strategy also clearly worked: in both experiments PD, the
rarest class, ended up the best identified. But the honest headline is
the domain gap. The model is genuinely strong inside the Kaggle
distribution and genuinely weak on an unseen cohort, and the failure is
specific---ADNI controls read as AD---rather than a general collapse,
which points at intensity and scanner differences rather than the
architecture. Read together with the calibration and conformal results,
the practical reading is that this system is a plausible pre-screening
aid within a known distribution, able to flag the cases it should not
decide alone, but not something to point at an arbitrary new scanner. It
assists a radiologist; it does not replace one.

There is a methodological point here we want to make plainly. Nothing
about the two experiments differs except the data scope and the split,
so the 25-point drop is neither noise nor a weaker model---it is simply
the cost of measuring honestly, and it is a cost the field mostly avoids
paying. We would argue that reporting it is more useful to a clinician
deciding whether to trust a tool than yet another near-perfect
single-dataset accuracy. The calibration and conformal numbers make that
argument concrete: because the model's confidence is well-behaved after
temperature scaling, and its prediction sets widen on the cases it finds
hard, a deployment could route the low-confidence or wide-set
scans---many of which are exactly the ADNI controls it struggles
with---to a specialist while clearing the confident majority
automatically. That selective behaviour, rather than a headline
accuracy, is what would let the system sit safely inside a real
workflow. It is also worth noting that the per-class story is stable
across both experiments: PD is the easiest class in each, and the
AD--Control pair is always the pivot, so the architecture's strengths
and weaknesses are consistent even as the absolute numbers move---the
split changes how well the model generalises, not what it finds hard.

\textbf{Limitations.} The model works on 2D axial slices and cannot see
volumetric biomarkers; ADNI's subject pool is small and has no PD;
Kaggle lacks subject IDs; and there is no prospective clinical
validation yet.

\section{Future Scope}
The obvious next step is to close the gap rather than merely report it.
Because the pipeline is already source-aware and supports
leave-one-dataset-out evaluation, domain-adversarial
training~\cite{b_dann} and ComBat-style intensity harmonization can be
measured directly against the 59.2\% ADNI baseline established here.
Beyond that we would like to move from 2D slices to 3D volumes with a
volumetric backbone~\cite{b_medicalnet}, extend the CFA idea to extra
modalities such as DTI and PET where paired data allows, add cohorts
like PPMI and OASIS, and push the uncertainty work toward deep ensembles
and concept-level explanations~\cite{b_tcav}.

Two of these matter more than the rest. Closing the Kaggle-to-ADNI gap
is the immediate priority, and because our splitter already keeps the
sources separate, a leave-one-dataset-out protocol gives a clean way to
check whether an adaptation method genuinely transfers rather than just
memorising a second distribution. Adding PD scans to the ADNI side---or
bringing in a third cohort such as PPMI that actually contains PD---would
let us test whether the near-perfect PD result survives outside Kaggle,
which the present data simply cannot tell us. Further out, prospective
validation, where the model reads scans as they arrive in a clinic and
radiologist agreement serves as the reference, is the only thing that
would move this work from a benchmark result toward something usable; we
treat every direction above as a step in that direction rather than an
end in itself.

A different lever on the PD shortage is to make more of it. Conditional
GANs~\cite{b_medigan} or diffusion models~\cite{b_diffusion_med} can
synthesise plausible PD-like slices, and while we are wary of training
on generated data---it is easy to reproduce the texture of a class
without reproducing its pathology---a carefully gated pipeline, in which
synthetic scans are screened by fidelity metrics and a neuroradiologist
before they ever enter training, could ease a scarcity that no amount of
resampling truly cures. In a similar spirit, self-supervised
pretraining~\cite{b_ssl_med} on the large pool of unlabelled MRI that
hospitals already hold would give the encoders a starting point grounded
in medical images rather than natural photographs. We suspect that would
help most exactly where the model is weakest today---on the unfamiliar
ADNI distribution---since the features learned that way are less tied to
any one scanner's intensity profile.

Two further directions look further out but matter for real deployment.
Federated learning would let several hospitals improve the model
jointly, especially on the minority PD class, without any of them
sharing raw patient scans, which sidesteps the privacy and governance
hurdles that usually stall multi-site medical AI before it starts. And
because neurodegeneration is a process rather than a snapshot, a
longitudinal variant trained on serial scans from the same patient could
watch for accelerating atrophy and flag change earlier than any
single-timepoint classifier can---which is where the clinical value of
imaging is highest, well before symptoms are obvious. Each of these
would also give the calibration and uncertainty machinery already in
place something more useful to act on, since a confidence estimate is
worth far more when it is attached to a trajectory than to a lone slice.

\section{Conclusion}
We did not set out to invent a new module; we set out to measure an
existing one honestly. Under the ordinary Kaggle-only, image-level
protocol HybridDualPathNet reaches 98.25\%---in line with the best
published results---but under a per-source patient-wise split across
Kaggle and ADNI the same network scores a defensible 73.4\%, with
well-behaved calibration and uncertainty and saliency that sits on
anatomy rather than artefacts. The number that matters is the 91.2\%
versus 59.2\% split between the two cohorts: an explicit, reproducible
measure of how far the model generalizes, and a gap the field usually
hides behind a single average. Reporting it, rather than the 98.25\% the
same network can produce under a leaky split, is the point of the paper.

More broadly, we think the two-experiment format is itself a small
contribution. Running one network under a leaky and a leak-free protocol,
and publishing both side by side, turns the usual single accuracy into a
measurement of how far that accuracy can be trusted---which is arguably
the number a clinician actually needs. The entire pipeline runs on a
single free-tier GPU with openly available data, so reproducing or
extending either experiment costs nothing but time. For a result whose
central claim is about honest evaluation, keeping the barrier to
reproduction that low feels like the right bar to set.

What we are ultimately left with is a model that is candid about its own
limits. Inside a known scanner distribution it performs at a level
competitive with the published state of the art, and it arrives with the
calibration, uncertainty, and saliency evidence a careful reviewer or
clinician would reasonably ask for before trusting it at all. Step
outside that distribution and it degrades---but, importantly, it says
so, both in the source-stratified accuracy and in the conformal sets
that widen on the cases it cannot resolve. A system that surfaces its
own uncertainty this plainly is, to our minds, closer to clinically
useful than one that quotes a higher single number and quietly buries
the rest. If this study nudges even a little toward reporting brain-MRI
classifiers under leakage-free, multi-cohort protocols as a matter of
course, we would consider that a more durable contribution than the
architecture itself.

\end{document}